\documentclass[11pt]{article}

\usepackage[preprint]{acl}
\usepackage{multirow}
\usepackage{times}
\usepackage{latexsym}
\usepackage{comment} 

\usepackage[T1]{fontenc}

\usepackage[utf8]{inputenc}

\usepackage{microtype}

\usepackage{inconsolata}

\usepackage{graphicx}

\usepackage{comment}
\usepackage{amsmath}
\usepackage{amssymb}
\usepackage{bbold}
\usepackage{booktabs}
\usepackage{xcolor}
\usepackage{float}
\usepackage{algorithm}
\usepackage[noend]{algpseudocode}
\usepackage{amsmath}   
\usepackage[most]{tcolorbox}

\usepackage{array}
\newcolumntype{L}[1]{>{\raggedright\arraybackslash}p{#1}}

\title{Training-Free Prompt Selection for Efficient Reinforcement Learning With Partial Rollouts}

\author{
  \textbf{Lei Jiang\textsuperscript{1}},
  \textbf{Honghui Bao\textsuperscript{1}},
  \textbf{Yue Zhou\textsuperscript{2}},
  \textbf{Zhengyao Gu\textsuperscript{1}},
\\
  \textbf{Langzhou He\textsuperscript{1}},
  \textbf{Natalie Parde\textsuperscript{1}}
\\
  \textsuperscript{1}University of Illinois Chicago,
  \textsuperscript{2}LLM Core and Agentic AI, CapitalOne
\\
  \texttt{
    {\{ljian43, hbao, zgu24, lhe24, parde\}}@uic.edu, yue.zhou@capitalone.com
  }
}

\begin{document}
\maketitle
\begin{abstract}
Reinforcement learning with verifiable rewards (RLVR) has emerged as a prominent paradigm for training reasoning models, but its computational cost makes efficient data selection increasingly important. Since training examples contribute unequally to learning, selecting a small, informative subset can offer a more efficient alternative to using the entire data pool. The training-free selection method~\cite{wu2026singlerollout}  estimates prompt value with hidden states from complete greedy rollouts. However, in our datasets, $26$--$32\%$ of these rollouts fail to terminate within the generation budget. We introduce LIFT, a training-free data selector that identifies impactful prompts using only partial reasoning rollouts. At its core, LIFT uses Partial Rollout Context Lift (PRCL), a score inspired by pointwise $\mathcal{V}$-information that accumulates token-level log-likelihood ratios between a question-conditioned context and a null context with the question removed. This enables prompt scoring from a short rollout prefix rather than a complete trajectory. Across Qwen2.5-Math-1.5B and DeepSeek-R1-Distill-1.5B, prefixes retaining at least 90\% and 95\% of the full-rollout PRCL score, respectively, require only 26\% and 4.7\% of the tokens generated by full rollouts. Compared with SHIFT, LIFT reduces selection cost by 3.7--9.1$\times$ while achieving competitive downstream RLVR performance, with the highest six-benchmark average in three of six settings. \footnote{Correspondence: \href{mailto:hbao@uic.edu}{hbao@uic.edu}}

\end{abstract}

\section{Introduction}
\begin{figure}[h]
\centering
\includegraphics[width=\columnwidth]{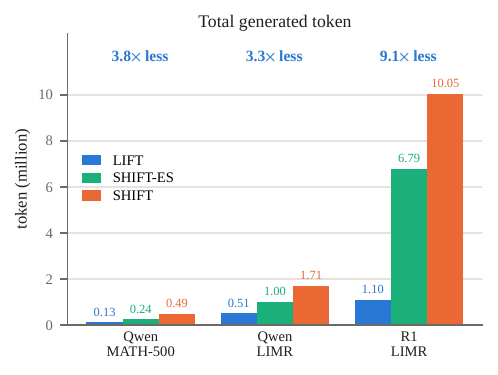}
\caption{Tokens generated to score each dataset and model by LIFT(ours), SHIFT-ES, and SHIFT.
}
\label{fig:main}
\end{figure}
\begin{figure*}[!t]
\centering
\includegraphics[width=\textwidth]{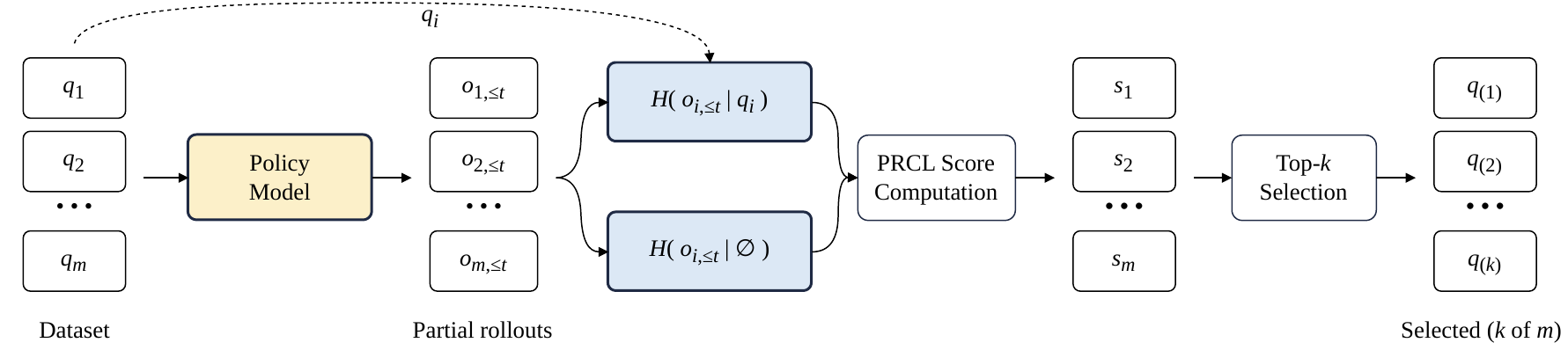}
\caption{Overview of LIFR. Given an input prompt, the LLM generates a greedy partial rollout, and the PRCL score is calculated from the entropy with and without the question. Then the prompts are ranked by it.
}
\label{fig:partial-main}
\end{figure*}
Reinforcement learning (RL), particularly with verifiable rewards,  has become a central paradigm for improving the reasoning capabilities of language models~\citep{lambert2024tulu3,shao2024deepseekmath,deepseekai2025r1,yu2025dapo}. However, generating multiple rollouts for each training example is computationally expensive because multiple responses are sampled for each prompt at every optimization step~\citep{zheng2025greso,yu2025dapo}.  Recent studies show that training on carefully selected subsets of examples can achieve performance comparable to training on the full data pool~\citep{li2025limr}, suggesting that identifying informative training examples can improve RL training efficiency.

Existing approaches to identifying impactful examples fall into two broad categories: training-dependent and training-free methods. Training-dependent methods leverage signals that emerge during RL training to identify or re-weight informative prompts~\citep{yu2025dapo,zheng2025greso,bae-etal-2026-online,sun2025improving,kim2026spendrollouts,li2025limr,wang2025oneshot,xia2024less,learnalign,yang2026gradalign}. However, these approaches typically require both access to verifiable rewards (which often entail ground-truth answers) and partial fine-tuning over a large candidate pool to obtain the signals needed for ranking. In contrast, training-free methods estimate the information that prompts can provide for RL by using different measures of prompt utility~\citep{sener2018active,li2026answerdivergence}. For example, SHIFT~\citep{wu2026singlerollout} uses hidden-state shifts from a single greedy rollout. By avoiding training-time prompt utility estimation and reward supervision, training-free methods provide a more general way to select prompts before RL training.

However, SHIFT's full trajectory requirement can be costly when greedy decoding becomes repetitive. On our two datasets, Math-500~\citep{math500} and LIMR~\citep{li2025limr}, 26--32\% of greedy trajectories fail to terminate within the generation budget, and most exhibit fixed $n$-gram repetition. We observed that such trajectories can spend a substantial number of tokens after the informative part of the reasoning has already been generated, suggesting that a complete trajectory may not be necessary to obtain an effective selection signal. We therefore ask: \emph{Under the training-free setting, can we efficiently select a high-quality subset of prompts with only a partial informative rollout?}

We answer this question by showing that effective prompt selection can be performed from partial rollouts. First, we introduce LIFT, a \textit{training-free} data selector that needs only a \textit{single, partial} reasoning trajectory.  Second, motivated by pointwise $\mathcal{V}$-information(PVI)~\cite{Etha2022Pvi}, which estimates example difficulty, we propose \textbf{Partial Rollout Context Lift} (PRCL).
PRCL measures how much the question changes the entropy of its partial reasoning trajectory compared with removing the question. We show that \textbf{partial rollouts suffice for effective training-free prompt selection}: short prefixes retain over 90\% of the selection signal of complete trajectories while requiring substantially fewer generated tokens. Third, the resulting subsets achieve \textbf{competitive downstream performance at substantially lower selection cost}.  Across models and selection budgets, PRCL-selected subsets achieve performance comparable to full-pool training and competitive with full-trajectory selection, while reducing selection cost by a factor of $3.7$--$9.1\times$. Additionally, inspired by the effectiveness of partial rollouts, we further adapt SHIFT with an early-stop rule that halts decoding at the onset of $n$-gram repetition. This cuts its selection cost by $1.5$ to $2.0\times$ at no loss in downstream accuracy in four settings.

\section{Related Work}

\paragraph{RL for LLM reasoning.}
RL was initially used in LLM post-training for preference alignment~\citep{ouyang2022instructgpt}, optimizing the policy against a learned reward model with proximal policy optimization (PPO)~\citep{schulman2017ppo}.
With automatically checkable answers, a verifier can replace the reward model; this approach is known as reinforcement learning with verifiable rewards (RLVR)~\citep{lambert2024tulu3}.
Group relative policy optimization (GRPO)~\citep{shao2024deepseekmath} can then drop PPO's value network, instead estimating advantages by normalizing rewards within a group of rollouts per prompt. GRPO has been used to train recent reasoning models (e.g., DeepSeek-R1~\citep{deepseekai2025r1}). Follow-up work refining the GRPO objective~\citep{yu2025dapo} decouples the lower and upper clipping bounds and discards prompts whose rollouts all receive the same reward.

\paragraph{Data Selection for LLM}
\textbf{Training-dependent methods} use signals that emerge during training to filter prompts or rollouts online~\citep{zheng2025greso,bae2026online,sun2025difficultytargeted}, reallocate rollout budgets across prompts~\citep{kim2026spendrollouts}, or identify informative prompts from reward curves, historical reward variance, or gradient similarity~\citep{li2025limr,wang2025oneshot,xia2024less,learnalign,yang2026gradalign}. E.g.GRESO~\citep{zheng2025greso} filters out the prompt with zero-variance based one the historical rewards before rollout. Analogous ``less is more~\citep{zhouLessismore2023}'' results have also been reported for supervised reasoning data~\citep{ye2025limo,muennighoff2025s1} by finding the minimal exemplars.

\textbf{Training-free methods}estimate prompt utility using signals extracted from the base model. Sener and Savarese~\citep{sener2018active} select representative examples by maximizing embedding-space coverage. Marion et al.~\citep{marion2023when} explore model uncertainty signals such as perplexity and entropy for assessing when a language model needs assistance. Li et al.~\citep{li2026answerdivergence} use the diversity of model-generated answers to guide instruction-data selection. SHIFT~\citep{wu2026singlerollout} estimates prompt utility from hidden-state shifts across a single greedy reasoning rollout and applies quality-weighted coreset selection to balance prompt utility and diversity. Our method falls into this category, using only token log-probabilities from a fixed-length rollout prefix.

\section{Methodology}

\begin{algorithm}[!t]
\caption{Training-free GRPO Data Selection via the PRCL}
\label{alg:prcl}
\begin{algorithmic}[1]
\State \textbf{Input:} Pool $\mathcal{U}=\{x_i\}_{i=1}^{N}$; budget $B$; base LM $\pi_\theta$; template $\mathcal{T}$;
       null context $\varnothing$; estimation fraction $\gamma$; target share $\alpha$; cap $K_{\max}$
\State \textbf{Output:} Selected subset $\mathcal{S}\subset\mathcal{U}$ with $|\mathcal{S}|=B$

\Function{Score}{$x$, $K$} 
  \State $y(x) \gets$ greedy decode from $\pi_\theta$ under $\mathcal{T}(x)$, at most $K$ tokens
  \For{$t = 1,\dots,|y(x)|$} 
    \State $s_t(x) \gets \log \pi_\theta\big(y_t \mid \mathcal{T}(x), y_{<t}\big)
                        - \log \pi_\theta\big(y_t \mid \mathcal{T}(\varnothing), y_{<t}\big)$
  \EndFor
  \State \Return $s(x)$
\EndFunction

\Statex \textbf{Phase 1: Estimate prefix length on a subset}
\State Draw $\mathcal{C}\subset\mathcal{U}$ at random, $|\mathcal{C}| = \lceil \gamma N \rceil$
\State $s(x) \gets$ \Call{Score}{$x$, $K_{\max}$} for every $x\in\mathcal{C}$
\State $\mathrm{PRCL}_K(x) \gets \sum_{t \le \min(K,\,|y(x)|)} s_t(x)$ for every $x\in\mathcal{C}$, $K \le K_{\max}$
\State $F(K) \gets \dfrac{\sum_{x\in\mathcal{C}} \mathrm{PRCL}_K(x)}{\sum_{x\in\mathcal{C}} \mathrm{PRCL}_{K_{\max}}(x)}$
\State $\hat{K} \gets \min\{K : F(K) \ge \alpha\}$

\Statex \textbf{Phase 2: Score the pool at $\hat{K}$}
\State $s(x) \gets$ \Call{Score}{$x$, $\hat{K}$} for every $x\in\mathcal{U}\setminus\mathcal{C}$
\State $\mathrm{PRCL}_{\hat{K}}(x) \gets \sum_{t \le \min(\hat{K}, |y(x)|)} s_t(x)$ for every $x\in\mathcal{U}$
\State $\mathcal{S} \gets \operatorname{Top}\text{-}B_{\,x\in\mathcal{U}}\; \mathrm{PRCL}_{\hat{K}}(x)$
\State \Return $\mathcal{S}$
\end{algorithmic}
\end{algorithm}

\paragraph{Task Formulation.}
Given an unlabeled prompt pool $\mathcal{U}$ and a selection budget $B$, our goal is to select an impactful subset $\mathcal{S}\subset\mathcal{U}$ of size $B$ for downstream RLVR training.  

We propose a training-free data selection method LIFT that uses only partial reasoning rollouts to identify impactful prompts. First, we randomly sample a subset of the pool to estimate an appropriate rollout prefix length. Then, we score the remaining prompts using this prefix length and select the top-$B$ prompts based on their PRCL scores.

\paragraph{Setup.}
Let $\mathcal{U}=\{x_i\}_{i=1}^{N}$ be an unlabeled pool of prompts and $B \ll N$ a selection budget.
Let $\pi_\theta$ be the initial policy and $\mathcal{T}(\cdot)$ the chat template used for GRPO training. For each prompt $x$, we generate a deterministic rollout $y(x)=(y_1,\dots,y_{L_x})$ using greedy decoding under $\mathcal{T}(x)$. We define a \emph{null context} $\mathcal{T}(\varnothing)$ by removing the user prompt while retaining the same template and system prompt.

\paragraph{Partial Rollout Context Lift (PRCL).}
Motivated by PVI~\cite{Etha2022Pvi}, which is used to the difficulty of a dataset in Classification Task, we create the PRCL metric.Given a context $c$ and a token sequence $y$, let
\begin{equation}
  H_\theta^{K}(y \mid c) \;=\; -\sum_{t=1}^{K} \log \pi_\theta\big(y_t \mid c,\, y_{<t}\big)
  \label{eq:H}
\end{equation}
be the entropy of the first $K$ tokens of $y$ under $\pi_\theta$ given $c$, computed by teacher forcing, where $K$ is the \emph{prefix length}.
Taking $c = \mathcal{T}(x)$ and $c = \mathcal{T}(\varnothing)$, and evaluating both on the same rollout $y = y(x)$, defines the Partial Rollout Context Lift, the amount by which the question lifts the likelihood of the models own partial rollout:
\begin{equation}
\begin{aligned}
  \mathrm{PRCL}_{K}(x) &= H_\theta^{K}\big(y \mid \mathcal{T}(\varnothing)\big) - H_\theta^{K}\big(y \mid \mathcal{T}(x)\big) \\[2pt]
  &= \sum_{t=1}^{K} s_t(x),
\end{aligned}
\label{eq:prcl}
\end{equation}
\begin{equation}
  s_t(x) = \log \frac{\pi_\theta(y_t \mid \mathcal{T}(x), y_{<t})}{\pi_\theta(y_t \mid \mathcal{T}(\varnothing), y_{<t})}.
  \label{eq:st}
\end{equation}
Each $s_t(x)$ is the log-ratio of the next-token probability with and without the question, so $\mathrm{PRCL}_K(x)$ measures how much the question changes the log-likelihood of the model's own continuation, accumulated over the prefix.

\paragraph{Estimate the prefix length.}
Equations~\eqref{eq:H}--\eqref{eq:st} hold for an arbitrary prefix length, so $K$ must be fixed before the pool can be ranked.  Truncating too early omits remaining influence exerted by the question, while extending too far delegates decoding effort to minimally contributing tokens.
We estimate $K$ from a randomly drawn subset of the pool, as shown by Phase 1 of Algorithm~\ref{alg:prcl}: a $\gamma$-fraction $\mathcal{C}$ is drawn and scored to a cap $K_{\max}$, from which we form the pooled cumulative share
\begin{equation}
\begin{aligned}
  F(K) &= \frac{\sum_{x\in\mathcal{C}} \mathrm{PRCL}_K(x)}{\sum_{x\in\mathcal{C}} \mathrm{PRCL}_{K_{\max}}(x)}, \\[2pt]
  \hat{K} &= \min\{K : F(K) \ge \alpha\},
\end{aligned}
\label{eq:mass}
\end{equation}
so that $\hat{K}$ is the shortest prefix at which the subset has accumulated a share $\alpha$ of its total score.
The target share $\alpha$ is set manually.
Phase 2 then scores the rest of the pool at $\hat{K}$ only.

\paragraph{Selection.}
Finally, we rank the pool by the score and keep the $B$ highest:
\begin{equation}
  \mathcal{S} \;=\; \operatorname*{Top\text{-}B}_{x \in \mathcal{U}} \; \mathrm{PRCL}_{\hat{K}}(x).
  \label{eq:select}
\end{equation}

The selected subset $\mathcal{S}$ is then used for GRPO training in place of the full pool. Selecting $\mathcal{S}$ requires only forward passes of $\pi_\theta$ over prefixes of at most $\hat{K}$ tokens (or $K_{\max}$ tokens for prompts in $\mathcal{C}$), without rewards, ground truth, or fine-tuning.
\section{Experiments}
\subsection{Setting}
\providecommand{\ms}[2]{#1{\scriptsize$_{\pm #2}$}}
\providecommand{\todo}[1]{\textcolor{red}{#1}}

\begin{table*}[!t]
\centering
\small
\setlength{\tabcolsep}{4pt}
\begin{tabular}{cclcccccccc}
\toprule
\multicolumn{3}{c}{\textbf{Training dataset}} & \multicolumn{7}{c}{\textbf{Evaluation benchmarks}} & \\
\cmidrule(lr){1-3} \cmidrule(lr){4-10}
Dataset & Budget & Method & Math500 & AIME24 & AMC & Gaokao & Miner. & Olymp. & Avg. & \# Sel. Tok. \\
\midrule
\multicolumn{11}{c}{\textit{Qwen2.5-Math-1.5B}} \\
\midrule
\multicolumn{3}{c}{Base model } & 33.5 & 10.8 & 29.4 & 18.0 & 6.2 & 21.3 & 19.9 & -- \\
\midrule
\multirow{11}{*}{\shortstack{MATH\\\scriptsize 350 train}}
  & \multirow{4}{*}{2\% (7)} & Random           & \ms{64.0}{1.2} & \ms{\textbf{10.9}}{1.8} & \ms{47.4}{2.3} & \ms{51.9}{1.1} & \ms{17.3}{1.1} & \ms{28.8}{0.6} & \ms{36.7}{0.6} & - \\
  &                      & SHIFT            & \ms{63.9}{2.6} & \ms{8.8}{4.2} & \ms{\textbf{48.8}}{2.4} & \ms{53.3}{0.7} & \ms{19.0}{0.9} & \ms{\textbf{28.9}}{0.7} & \ms{\textbf{37.1}}{0.9} & 0.49M \\
  &                      & SHIFT-ES         & \ms{\textbf{65.7}}{0.9} & \ms{6.9}{2.1} & \ms{42.3}{1.7} & \ms{\textbf{55.4}}{1.6} & \ms{\textbf{20.3}}{0.9} & \ms{28.2}{0.7} & \ms{36.5}{0.6} & 0.24M \\
  &                      & LIFT (ours)  & \ms{63.8}{0.9} & \ms{7.7}{2.7} & \ms{46.2}{1.8} & \ms{53.1}{0.4} & \ms{17.4}{1.1} & \ms{27.6}{0.7} & \ms{36.0}{1.0} & 0.13M \\
\cmidrule(lr){2-11}
  & \multirow{4}{*}{4\% (14)} & Random           & \ms{65.7}{1.3} & \ms{\textbf{11.5}}{2.7} & \ms{\textbf{49.8}}{1.3} & \ms{53.1}{1.2} & \ms{18.4}{0.9} & \ms{29.5}{0.7} & \ms{38.0}{0.8} & - \\
  &                      & SHIFT            & \ms{67.5}{2.3} & \ms{9.4}{2.4} & \ms{47.5}{3.9} & \ms{55.3}{3.2} & \ms{20.4}{2.3} & \ms{30.5}{2.1} & \ms{\textbf{38.4}}{2.1} & 0.49M \\
  &                      & SHIFT-ES         & \ms{67.7}{2.6} & \ms{8.8}{3.7} & \ms{45.2}{0.9} & \ms{\textbf{56.7}}{2.1} & \ms{\textbf{21.1}}{1.3} & \ms{30.5}{1.0} & \ms{38.3}{0.6} & 0.24M \\
  &                      & LIFT (ours)  & \ms{\textbf{69.1}}{3.3} & \ms{10.0}{1.7} & \ms{44.8}{1.4} & \ms{56.6}{3.0} & \ms{18.9}{2.0} & \ms{\textbf{30.6}}{1.6} & \ms{\textbf{38.4}}{1.8} & 0.13M \\
\cmidrule(lr){2-11}
  & \multirow{2}{*}{100\%} & Full data      & \ms{69.0}{0.5} & \ms{10.6}{3.6} & \ms{50.2}{2.6} & \ms{55.4}{1.5} & \ms{19.7}{0.8} & \ms{31.6}{0.9} & \ms{39.4}{0.5} & -- \\
  &                      & GRESO           & \ms{71.2}{1.0} & \ms{12.1}{2.4} & \ms{48.3}{1.6} & \ms{57.3}{2.1} & \ms{20.3}{1.0} & \ms{31.7}{0.6} & \ms{40.1}{0.6} & -- \\
\midrule
\multirow{11}{*}{\shortstack{LIMR\\\scriptsize 1{,}389}}
  & \multirow{4}{*}{2\% (28)} & Random           & \ms{67.6}{1.5} & \ms{13.2}{2.9} & \ms{48.3}{2.9} & \ms{\textbf{55.6}}{1.5} & \ms{20.4}{0.9} & \ms{29.9}{1.0} & \ms{39.2}{0.9} & - \\
  &                      & SHIFT            & \ms{\textbf{69.2}}{0.1} & \ms{11.9}{2.2} & \ms{\textbf{50.0}}{2.2} & \ms{55.5}{0.6} & \ms{21.4}{0.8} & \ms{\textbf{31.7}}{0.4} & \ms{39.9}{0.5} & 1.71M \\
  &                      & SHIFT-ES & \ms{68.5}{0.5} & \ms{10.8}{0.8} & \ms{\textbf{50.0}}{2.9} & \ms{55.0}{0.3} & \ms{21.0}{1.0} & \ms{30.5}{0.2} & \ms{39.3}{0.4} & 1.00M \\
  &                      & LIFT (ours)  & \ms{69.2}{0.6} & \ms{\textbf{13.8}}{1.4} & \ms{49.4}{1.5} & \ms{54.2}{0.6} & \ms{\textbf{22.6}}{0.8} & \ms{31.4}{0.4} & \ms{\textbf{40.1}}{0.3} & 0.51M \\
\cmidrule(lr){2-11}
  & \multirow{4}{*}{4\% (56)} & Random           & \ms{66.9}{0.8} & \ms{\textbf{14.3}}{2.0} & \ms{50.1}{1.0} & \ms{54.2}{1.3} & \ms{20.1}{0.9} & \ms{30.2}{0.4} & \ms{39.3}{0.3} & - \\
  &                      & SHIFT            & \ms{69.0}{0.9} & \ms{11.9}{1.8} & \ms{46.9}{1.5} & \ms{54.8}{0.8} & \ms{21.9}{0.9} & \ms{31.2}{0.6} & \ms{39.3}{0.7} & 1.71M \\
  &                      & SHIFT-ES         & \ms{68.8}{0.5} & \ms{12.7}{2.8} & \ms{49.8}{2.2} & \ms{54.7}{0.1} & \ms{\textbf{22.2}}{0.6} & \ms{31.7}{0.5} & \ms{40.0}{0.5} & 1.00M \\
  &                      & LIFT (ours)  & \ms{\textbf{69.9}}{1.1} & \ms{12.1}{2.1} & \ms{\textbf{51.9}}{3.1} & \ms{\textbf{56.1}}{1.0} & \ms{22.0}{0.2} & \ms{\textbf{32.0}}{0.5} & \ms{\textbf{40.6}}{0.7} & 0.51M \\
\cmidrule(lr){2-11}
  & \multirow{2}{*}{100\%} & Full data      & \ms{69.3}{0.5} & \ms{14.0}{2.0} & \ms{48.9}{2.6} & \ms{55.2}{0.5} & \ms{21.3}{1.0} & \ms{31.8}{0.8} & \ms{40.1}{0.4} & -- \\
  &                      & GRESO           & \ms{70.0}{1.0} & \ms{14.6}{2.6} & \ms{52.3}{0.6} & \ms{56.2}{0.2} & \ms{22.0}{0.4} & \ms{32.2}{0.2} & \ms{41.2}{0.6} & -- \\
\midrule
\multicolumn{11}{c}{\textit{DeepSeek-R1-Distill-Qwen-1.5B}} \\
\midrule
\multicolumn{3}{c}{Base model} & 83.1 & 26.7 & 66.9 & 73.0 & 27.8 & 44.3 & 53.6 & -- \\
\midrule
\multirow{11}{*}{\shortstack{LIMR\\\scriptsize 1{,}389}}
  & \multirow{4}{*}{2\% (28)} & Random           & \ms{84.7}{0.7} & \ms{31.9}{4.7} & \ms{73.0}{1.4} & \ms{75.1}{0.3} & \ms{28.9}{0.4} & \ms{47.1}{0.2} & \ms{56.8}{1.0} & - \\
  &                      & SHIFT            & \ms{84.6}{0.5} & \ms{30.4}{3.1} & \ms{75.0}{3.8} & \ms{75.3}{0.6} & \ms{29.3}{0.6} & \ms{46.8}{0.7} & \ms{56.9}{0.6} & 10.05M \\
  &                      & SHIFT-ES & \ms{\textbf{85.5}}{0.4} & \ms{\textbf{31.9}}{2.9} & \ms{\textbf{75.2}}{1.4} & \ms{\textbf{76.4}}{0.1} & \ms{\textbf{30.1}}{0.6} & \ms{\textbf{49.3}}{0.6} & \ms{\textbf{58.1}}{0.4} & 6.79M \\
  &                      & LIFT (ours)  & \ms{84.2}{0.1} & \ms{31.9}{4.2} & \ms{74.1}{1.2} & \ms{75.6}{0.6} & \ms{29.5}{0.8} & \ms{46.6}{0.6} & \ms{57.0}{1.0} & 1.10M \\
\cmidrule(lr){2-11}
  & \multirow{4}{*}{4\% (56)} & Random           & \ms{84.6}{0.6} & \ms{30.4}{2.2} & \ms{73.4}{1.9} & \ms{\textbf{75.8}}{0.8} & \ms{29.2}{1.2} & \ms{47.2}{0.7} & \ms{56.8}{0.9} & - \\
  &                      & SHIFT            & \ms{84.4}{0.4} & \ms{30.2}{2.6} & \ms{74.5}{2.4} & \ms{75.7}{1.0} & \ms{\textbf{29.8}}{1.1} & \ms{47.9}{0.3} & \ms{57.1}{0.7} & 10.05M \\
  &                      & SHIFT-ES & \ms{83.8}{1.0} & \ms{31.4}{2.7} & \ms{\textbf{75.0}}{1.1} & \ms{74.5}{1.2} & \ms{29.1}{0.4} & \ms{\textbf{48.4}}{1.1} & \ms{57.0}{0.7} & 6.79M \\
  &                      & LIFT (ours)  & \ms{\textbf{84.7}}{0.4} & \ms{\textbf{32.5}}{2.4} & \ms{74.8}{0.6} & \ms{75.5}{0.2} & \ms{29.0}{0.9} & \ms{47.3}{0.5} & \ms{\textbf{57.3}}{0.5} & 1.10M \\
\cmidrule(lr){2-11}
  & \multirow{2}{*}{100\%} & Full data      & \ms{83.9}{0.6} & \ms{30.0}{4.4} & \ms{74.1}{2.9} & \ms{75.1}{0.4} & \ms{29.6}{0.8} & \ms{47.1}{0.9} & \ms{56.6}{0.8} & -- \\
  &                      & GRESO           & \ms{84.2}{1.2} & \ms{28.3}{7.1} & \ms{76.9}{2.7} & \ms{75.7}{1.2} & \ms{29.9}{1.5} & \ms{47.9}{2.0} & \ms{57.2}{1.7} & -- \\
\bottomrule
\end{tabular}
\caption{Accuracy (\%) on math reasoning benchmarks. The left three columns identify the \emph{training} dataset, budget, and selection method; the remaining columns are \emph{evaluation} benchmarks, their average, and the number of tokens generated for selection (\# Sel.~Tok.). }
\label{tab:main}
\end{table*}

\paragraph{Models and datasets.}
due to xx
 We train two models, Qwen2.5-Math-1.5B~\citep{yang2024qwen2} and DeepSeek-R1-Distill-Qwen-1.5B~\citep{deepseekai2025r1}, using subsets selected with budgets of 2\% and 4\% of the respective datasets. For Qwen2.5-Math-1.5B, we select 7 and 14 prompts from MATH-500~\citep{math500} and 28 and 56 prompts from LIMR~\citep{li2025limr}, respectively. DeepSeek-R1-Distill-Qwen-1.5B is trained only on LIMR using the same budgets.\footnote{We exclude DeepSeek-R1-Distill-Qwen-1.5B from training on MATH-500 because its initial accuracy on MATH-500 is already 83.1\%, leaving limited headroom for comparing selection methods.} MATH-500 is split into 350 training problems and 150 held-out evaluation problems. 
 We evaluate the trained models on six mathematical reasoning benchmarks: MATH-500, AIME24~\cite{aime24}, AMC23~\citep{AMC}, Gaokao~\citep{Zhang2023gaokao}, Minerva Math~\citep{Minerva}, and OlympiadBench~\citep{he2024olympiadbench}. For each model, prompt selection is performed by the same model used to initialize RL training, relying exclusively on problem text without access to answers or rewards.
 
 All runs use two NVIDIA GH200 96 GB GPUs, and a single seed takes from $1.0$ hour for Qwen2.5-Math-1.5B on the MATH training split to $13.5$ hours for DeepSeek-R1-Distill on LIMR.

\paragraph{Training and evaluation.}
Each model is trained on selected subsets with GRPO~\citep{shao2024deepseekmath} in \texttt{verl} \cite{10.1145/3689031.3696075}, with identical hyperparameters for every selection method: learning rate $2\times10^{-6}$, eight rollouts per prompt at temperature 0.7, and a low-variance KL loss with coefficient $10^{-3}$.
Qwen2.5-Math-1.5B uses a 4{,}096-token context (at most 1{,}408 prompt and 2{,}688 response tokens), and DeepSeek-R1-Distill uses a 12{,}288-token response budget, matching the training length for each model.
We train for 270 steps and evaluate the final checkpoint with accuracy averaged over four samples per problem (avg@4) at temperature 0.6 and top-$p$ 0.95, graded with MathRuler~\citep{mathruler}. We report the mean and standard deviation over four training seeds.

\paragraph{Comparison methods.}
We evaluate performance relative to a variety of baselines and alternative approaches.  \textbf{Random} draws a subset uniformly from the dataset, each trained with four seeds.
\textbf{PRCL} scores each prompt by $\mathrm{PRCL}_{\hat{K}}$ over a prefix of its greedy rollout and keeps the top $B$ prompts, with $\hat{K}$ estimated from a random 10\% of the dataset at a target share $\alpha$.
We set $\alpha = 0.90$ for Qwen2.5-Math-1.5B and $\alpha = 0.95$ for DeepSeek-R1-Distill;
\textbf{SHIFT}~\citep{wu2026singlerollout} ranks prompts by the hidden-state shift of a single greedy rollout and selects with quality-weighted farthest-first coverage.  While SHIFT reads the hidden state at the final token of the rollout (its anchor), \textbf{SHIFT-ES} reads it at the last non-repeated token and stops decoding at the first repetition.
The signal and the selection rule are otherwise unchanged, thus isolating the anchor.
\textbf{Full data} trains on the entire dataset for the same number of steps. We compare all subsets under the same training setting.
And finally, \textbf{GRESO}~\citep{zheng2025greso} filters during training instead.  It predicts from each prompt's reward history which groups will be zero-variance, and accordingly skips their rollouts.

\subsection{Results}

\begin{figure}[h]
\centering
\includegraphics[width=\columnwidth]{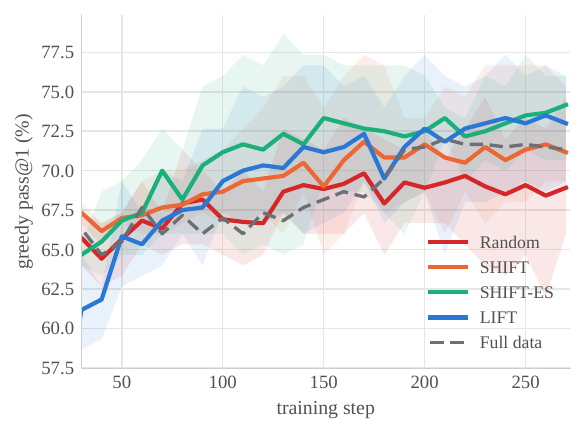}
\caption{Held-out accuracy during training for Qwen2.5-Math-1.5B under 4\% training setting, evaluated every 10 steps.}
\label{fig:curves}
\end{figure}

As an example, Figure~\ref{fig:curves} traces held-out accuracy through training at the $4\%$ budget on MATH. The results yield several key findings, elaborated as follows.

\begin{table}[h]
\centering
\footnotesize
\setlength{\tabcolsep}{6pt}
\begin{tabular}{llcc}
\toprule
Scoring model & Dataset & 2\% & 4\% \\
\midrule
\multirow{2}{*}{Qwen2.5-Math}
  & MATH-500 & 2 / 7   & 5 / 14  \\
  & LIMR     & 13 / 28 & 30 / 56 \\
\midrule
R1-Distill
  & LIMR     & 3 / 28  & 5 / 56  \\
\bottomrule
\end{tabular}
\caption{Prompts common to SHIFT and SHIFT-ES selections.}
\label{tab:anchor-overlap}
\end{table}
\paragraph{Repetition has limited impact on SHIFT's selection quality.} We shows the 
Moving SHIFT's end anchor before repeated spans substantially changes the selected prompts, as shown in Table~\ref{tab:anchor-overlap}, with Qwen2.5-Math retaining 29--54\% overlap and DeepSeek-R1-Distill only 5--11\%. Despite these changes, downstream accuracy remains similar.  Across six settings, SHIFT-ES improves the average by only 0.1 points over SHIFT, within seed standard deviations of 0.4--2.1. Meanwhile, scoring LIMR with DeepSeek-R1-Distill requires 10.05M generated tokens, 61\% spent on non-terminating rollouts, highlighting the decoding cost of full rollouts.

\paragraph{LIFT is competitive across models and datasets while saving scoring cost.}
LIFT has the highest six-benchmark average in three of the six settings, all of them on LIMR.
Qwen2.5-Math reaches $40.1$ at the 2\% budget and $40.6$ at 4\%, against $39.9$ and $40.0$ for the best baseline in each, and R1-Distill reaches $57.3$ at 4\% against $57.1$.
In the remaining three settings LIFT is second at the R1-Distill 2\% budget with $57.0$ behind SHIFT-ES at $58.1$, level with SHIFT at MATH-500 4\% at $38.4$, and last at MATH-500 2\% with $36.0$ against $37.1$.
That last setting selects seven prompts in total, which hints that LIFT may need a budget of more than a few prompts to be effective.

Against training on all of the data, LIFT-selected subsets lose no accuracy on LIMR.
Qwen2.5-Math achieves accuracies of $40.1$ and $40.6$ with 2\% and 4\% of the 1{,}389 prompts, against $40.1$ for the full dataset, and R1-Distill achieves accuracies of $57.0$ and $57.3$ compared to $56.6$.
On the 350-problem MATH-500 training set, performance is observed to be $36.0$ and $38.4$ accuracy compared to $39.4$.
LIFT scoring costs $0.13$M generated tokens on MATH-500 and $0.51$M and $1.10$M on LIMR, each including the Phase~1 estimation pass over a random $10\%$ of the dataset, compared to $0.49$M, $1.71$M, and $10.05$M for SHIFT. LIFT therefore generates $3.3$ to $9.1\times$ fewer tokens than SHIFT and $1.9$ to $6.2\times$ fewer than SHIFT-ES.

\paragraph{An online rollout selection baseline.}
GRESO filters during training rather than selecting data, training on all 350 or 1{,}389 prompts and skipping rollouts for groups it predicts will be zero-variance.  It achieves accuracies of $40.1$ and $41.2$ on the two Qwen2.5-Math settings, relative to $39.4$ and $40.1$ for full dataset training. It achieves an accuracy of $57.2$ on R1, compared to $56.6$ for full dataset training. We include it as a compatible reference point from the other family of methods rather than as a competing selector.

\section{Additional Studies}

\providecommand{\todo}[1]{\textcolor{red}{#1}}

\subsection{PRCL Distribution}
\label{sec:token-prcl}

\begin{figure}[!h]
\centering
\includegraphics[width=\columnwidth]{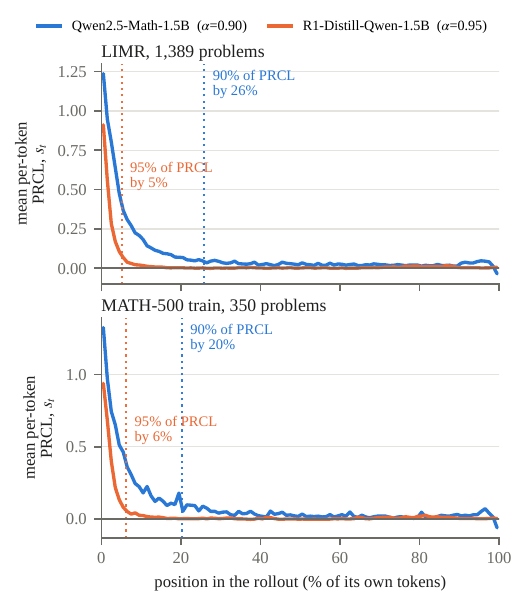}
\caption{Prompt influence by position along greedy rollouts, shown as mean per-token $s_t$ in 1\% bins of each rollout's own length, with every rollout weighted equally, on LIMR (top) and the MATH-500 training set (bottom).}
\label{fig:token-position}
\end{figure}

To understand why a prefix of the rollout can be sufficient for impactful prompt selection, we analyze the distribution of PRCL across models and datasets to see how it changes as the rollout length grows. We decode one full greedy rollout per prompt for all four model and dataset combinations, obtain the per-token log-ratio $s_t(x)$ of Eq.~\eqref{eq:st}, and average it by position.  Our findings are summarized in Figure~\ref{fig:token-position}.

\paragraph{Score distribution.}
PRCL rises steeply over the opening tokens and then flattens.
The median rollout reaches its target share within roughly the first quarter of the answer for Qwen2.5-Math, at 25.9\% on LIMR and 20.3\% on MATH-500, and within the first few percentage points for R1-Distill, at 5.1\% and 6.3\%.
Beyond those points the curve is almost flat, so most of the selection signal lies in an early prefix, therefore enabling prefix-based scoring.

\paragraph{Effect of repetition.}
The flat tail consists largely of repeated tokens.
About a fifth of greedy rollouts end in a verbatim loop (21.3\% for Qwen and 22.5\% for R1-Distill), and those loops account for 41.8\% and 32.4\% of all generated tokens while conveying very little of the score. Qwen's mean $|s_t|$ falls from 0.0998 before loop onset to 0.0012 after, since inside a loop both contexts predict each repeated token from the rollout itself.  Conditioning on the problem has almost no effect.

\paragraph{Model dependence.}
The two panels of Figure~\ref{fig:token-position} (i.e., the two datasets) differ far less than the two curves within each panel (i.e., the two models).
Qwen2.5-Math reaches its target share at 25.9\% and 20.3\% of the answer, compared to R1-Distill's 5.1\% and 6.3\%, and 55.5\% and 50.4\% of its tokens carry $s_t > 0$, compared to 25.8\% and 25.6\%. In both cases, the gap between the models is several times the gap between the two datasets of one model.
The tightness of the signal concentration therefore depends on the model's generation behavior rather than on the problems it is given, motivating the prefix length estimation per model that we analyze in Section~\ref{sec:estimation}.

\subsection{Characteristics of Selected Subsets}
\label{sec:subsets}

Different prefix lengths can change the PRCL scores and thus the relative ranking of prompts; This may lead to different selected subsets. To analyze this effect, we vary $\alpha$ to obtain different prefix lengths across all 1,389 LIMR prompts and measure the overlap between the resulting top-$B$ subsets and the subset selected using full trajectories. To characterize the subsets selected by different methods, we compare the solve rate $p(x)$ of their selected subset, as solve rate is considered an important learning signal in GRPO~\citep{bae-etal-2026-online, shao2024deepseekmath}.

\paragraph{PRCL ranking robustness.}
\begin{table}[h]
\centering
\footnotesize
\setlength{\tabcolsep}{5pt}
\begin{tabular}{llcccc}
\toprule
 & & \multicolumn{4}{c}{Target Share $\alpha$} \\
\cmidrule(lr){3-6}
Model / Dataset & & .85 & .90 & .95 & .99 \\
\midrule
\multirow{3}{*}{Qwen / MATH-500}
  & $\hat{K}$   & 160 & 260 & 418 & 752 \\
  & $\cap_{7}$  & 6   & 7   & 7   & 7   \\
  & $\cap_{14}$ & 10  & 12  & 12  & 12  \\
\midrule
\multirow{3}{*}{Qwen / LIMR}
  & $\hat{K}$   & 208 & 340 & 520 & 952 \\
  & $\cap_{28}$ & 13  & 18  & 21  & 26  \\
  & $\cap_{56}$ & 29  & 36  & 43  & 52  \\
\midrule
\multirow{3}{*}{R1 / LIMR}
  & $\hat{K}$   & 120 & 173 & 299 & 552 \\
  & $\cap_{28}$ & 10  & 11  & 14  & 17  \\
  & $\cap_{56}$ & 23  & 23  & 30  & 36  \\
\bottomrule
\end{tabular}
\caption{Target share versus selected set. $\hat{K}$: the prefix length in tokens at which the whole dataset has accumulated a share $\alpha$ of its PRCL. $\cap_B$: prompts shared by the top-$B$ set at $\hat{K}$ and the top-$B$ set of the full rollout, at the 2\% and 4\% budgets.}
\label{tab:rank-k}
\end{table}
Table~\ref{tab:rank-k} sweeps the target share $\alpha$ and reports how many prompts in the full-trajectory selection remain selected at the prefix length each value yields.
The sweep changes $\hat{K}$ by a factor of four to five but changes the selection far less. The 2\% set on MATH-500 is already the full-trajectory set from $\alpha = 0.90$, and on LIMR its overlap grows by 13 prompts for Qwen2.5-Math and 7 for R1-Distill over the whole range.
R1-Distill agrees least at every $\alpha$, as Section~\ref{sec:token-prcl} predicts, since its order keeps shifting through the long reasoning that follows the early mass.

\paragraph{Subset characteristics.}
\label{sec:solve-rate}

To characterize the selection of each method, we estimate every LIMR prompt's solve rate $p(x)$~\cite{zhou-etal-2024-paraphrase}, defined as the fraction of correctly answered rollouts among 16 samples at temperature 0.7.
GRPO produces a gradient on $x$ only when a group of rollouts contains both correct and incorrect answers, so prompts with $p(x) \in \{0,1\}$ produce no signal in expectation. Since solve rates are unavailable to training-free selectors, we use them post hoc to characterize the selected subsets in terms of prompt difficulty.
\begin{figure*}[!t]
\centering
\includegraphics[width=\textwidth]{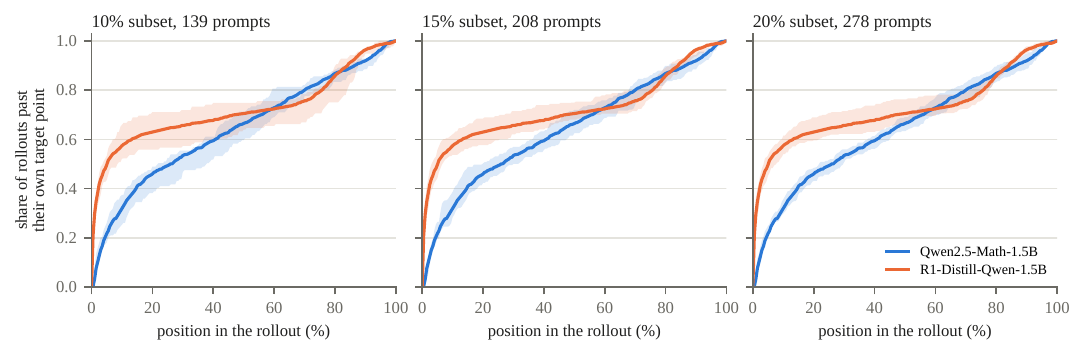}
\caption{Random subset recovery of the whole dataset's distribution, at three subset sizes. Each panel plots the share of LIMR rollouts that have collected their target share $\alpha$ of their own PRCL by a given position.  The line is the whole 1{,}389-prompt dataset, and the band spans ten independent random draws.}
\label{fig:subset-recovery}
\end{figure*}
\begin{table}[h]
\centering
\footnotesize
\setlength{\tabcolsep}{4pt}
\begin{tabular}{llcccccc}
\toprule
 & & \multicolumn{3}{c}{2\% (28)} & \multicolumn{3}{c}{4\% (56)} \\
\cmidrule(lr){3-5}\cmidrule(lr){6-8}
Model & Method & $\bar{p}$ & $p{=}0$ & $p{\le}.25$ & $\bar{p}$ & $p{=}0$ & $p{\le}.25$ \\
\midrule
\multirow{3}{*}{Qwen}
  & Random & .324 & 4  & 14 & .275 & 12 & 31 \\
  & SHIFT  & .174 & 5  & 24 & .206 & 8  & 42 \\
  & PRCL   & .114 & 9  & 23 & .142 & 16 & 45 \\
\midrule
\multirow{3}{*}{R1}
  & Random & .442 & 7  & 13 & .472 & 15 & 23 \\
  & SHIFT  & .163 & 18 & 22 & .081 & 46 & 50 \\
  & PRCL   & .234 & 16 & 19 & .277 & 26 & 36 \\
\bottomrule
\end{tabular}
\caption{Solve rate of the selected LIMR subsets, each scored by the model that selects and trains on it: subset mean $\bar{p}$ and the number of prompts with $p = 0$ and with $p \le 0.25$. $p$ is estimated from 16 sampled rollouts per prompt at temperature 0.7 for both models.}
\label{tab:solve-rate}
\end{table}

As shown in Table~\ref{tab:solve-rate} selectors move toward difficult prompts, even though by different margins for the two models.
For Qwen2.5-Math, PRCL is the more concentrated, with a subset mean of $\bar{p} = 0.11$ at the 2\% budget against 0.17 for SHIFT and 0.33 for the dataset, and 23 of its 28 prompts at $p \le 0.25$.
For R1-Distill the ordering reverses: SHIFT reaches $\bar{p} = 0.16$ at 2\% and 0.08 at 4\%, where 46 of its 56 prompts are never solved, against 0.23 and 0.28 for PRCL.
Both therefore retain many prompts the policy never solves, which contribute no gradient until it solves them at least once.

\subsection{Random Subset Ratio for Prefix-Length Estimation}
\label{sec:estimation}

Since estimating the prefix length from a random subset introduces sampling variability, we investigate how the subset size affects the reliability of the estimate. Specifically, we estimate the full-dataset PRCL distribution using 10\%, 15\%, and 20\% of the data, and compare each estimate with the full-dataset distribution and the corresponding prefix length it yields. 

\begin{table}[h]
\centering
\footnotesize
\setlength{\tabcolsep}{4pt}
\begin{tabular}{lcccc}
\toprule
 & \multicolumn{2}{c}{Qwen2.5-Math} & \multicolumn{2}{c}{R1-Distill} \\
\cmidrule(lr){2-3}\cmidrule(lr){4-5}
$\alpha$ & med. & range & med. & range \\
\midrule
$0.80$ & 128 & 112--208 & 96  & 64--144 \\
$0.90$ & 336 & 224--496 & 176 & 96--448 \\
$0.95$ & 528 & 416--736 & 288 & 128--832 \\
\bottomrule
\end{tabular}
\caption{Estimated prefix length $\hat{K}$ in tokens at three target shares, on greedy LIMR rollouts: the median and the full minimum to maximum range over 300 random 10\% subsets.}
\label{tab:subset-k}
\end{table}
\paragraph{Estimation error.} 
Table~\ref{tab:subset-k} reports $\hat{K}$ in tokens over 300 random $\gamma = 0.1$ draws, 139 prompts each.
At the model-specific settings, Qwen2.5-Math with $\alpha = 0.90$ yields a token-length range of 224–496, with a median of 336, while R1-Distill with $\alpha = 0.95$ yields a range of 128–832, with a median of 288. The upper bounds are influenced by only a few random draws, making the median a more representative reference.

As shown in Figure~\ref{fig:subset-recovery},
the largest gap between a subset curve and the whole dataset curve falls from 0.096 to 0.055 for Qwen as $\gamma$ goes from 0.1 to 0.2, and from 0.119 to 0.066 for R1-Distill. Estimation error improves with the larger size of the random subset. 
The 350-problem MATH-500 training set is a harder basis for estimation, since 10\% of it is 35 prompts. The Qwen spread widens to 16.8--35.4\%, and the worst gap widens to 0.19.

\section{Conclusion}
We investigated whether training-free prompt selection for RLVR requires decoding complete greedy rollouts and found that short prefixes can provide sufficient information for effective selection. 
We introduce LIFT, a training-free RLVR data selector that estimates prompt value from partial reasoning rollouts using a token-level information score. The core of LIFT is the Partial Rollout Context Lift (PRCL), which measures how the question context changes the log-likelihood of the opening tokens of the model's own greedy rollout relative to a null context with the question removed. 

PRCL's cumulative score allows multiple prefix lengths to be evaluated from a single decode. Moreover, the prefix length can be estimated using a random $10\%$ of the candidate pool rather than fixed in advance. By using partial rollouts, PRCL reduces dataset-level scoring cost by $3.7$--$9.1\times$ relative to SHIFT. Across six benchmark and budget settings, it achieves the highest average downstream performance in three settings while remaining competitive with full trajectory selection and full pool training.

\section{Limitations}

\paragraph{Choosing the prefix length.}
Our prefix-length rule identifies the point at which a random $\gamma$-fraction of the dataset has accumulated a target share $\alpha$ of its score. Although this replaces an absolute token-count heuristic with a score-based criterion, the target share $\alpha$ remains model-dependent and must be calibrated for each model. We set $\alpha=0.90$ for Qwen2.5-Math and $\alpha=0.95$ for DeepSeek-R1-Distill based on their score accumulation curves. The estimate is also sensitive to the sampled subset: at $\gamma=0.1$, $\hat{K}$ ranges from 224 to 496 tokens for Qwen2.5-Math and from 128 to 832 for DeepSeek-R1-Distill. Although the median matches the whole-dataset estimate, increasing $\gamma$ is needed to reduce variability, leaving a more robust, model-agnostic prefix-length estimation for future work.

\paragraph{Scope of the evaluation.}
We evaluate two 1.5B models on two mathematics datasets at selection budgets of 2\% and 4\%.
The smallest setting selects seven prompts, where the differences between methods fall within the spread across training seeds.
Whether the early concentration of PRCL holds at larger model scales, on other reasoning domains, or at larger budgets remains untested.


\bibliography{custom}

@misc{schulman2017ppo,
      title={Proximal Policy Optimization Algorithms}, 
      author={John Schulman and Filip Wolski and Prafulla Dhariwal and Alec Radford and Oleg Klimov},
      year={2017},
      eprint={1707.06347},
      archivePrefix={arXiv},
      primaryClass={cs.LG},
      url={https://arxiv.org/abs/1707.06347}, 
}

@inproceedings{
sun2025improving,
title={Improving Data Efficiency for {LLM} Reinforcement Fine-tuning Through Difficulty-targeted Online Data Selection and Rollout Replay},
author={Yifan Sun and Jingyan Shen and Yibin Wang and Tianyu Chen and Zhendong Wang and Mingyuan Zhou and Huan Zhang},
booktitle={The Thirty-ninth Annual Conference on Neural Information Processing Systems},
year={2025},
url={https://openreview.net/forum?id=uwUkETPIJN}
}

@inproceedings{ouyang2022instructgpt,
 author = {Ouyang, Long and Wu, Jeffrey and Jiang, Xu and Almeida, Diogo and Wainwright, Carroll and Mishkin, Pamela and Zhang, Chong and Agarwal, Sandhini and Slama, Katarina and Ray, Alex and Schulman, John and Hilton, Jacob and Kelton, Fraser and Miller, Luke and Simens, Maddie and Askell, Amanda and Welinder, Peter and Christiano, Paul F and Leike, Jan and Lowe, Ryan},
 booktitle = {Advances in Neural Information Processing Systems},
 doi = {10.52202/068431-2011},
 editor = {S. Koyejo and S. Mohamed and A. Agarwal and D. Belgrave and K. Cho and A. Oh},
 pages = {27730--27744},
 publisher = {Curran Associates, Inc.},
 title = {Training language models to follow instructions with human feedback},
 url = {https://proceedings.neurips.cc/paper_files/paper/2022/file/b1efde53be364a73914f58805a001731-Paper-Conference.pdf},
 volume = {35},
 year = {2022}
}

@misc{lambert2024tulu3,
      title={Tulu 3: Pushing Frontiers in Open Language Model Post-Training}, 
      author={Nathan Lambert and Jacob Morrison and Valentina Pyatkin and Shengyi Huang and Hamish Ivison and Faeze Brahman and Lester James V. Miranda and Alisa Liu and Nouha Dziri and Shane Lyu and Yuling Gu and Saumya Malik and Victoria Graf and Jena D. Hwang and Jiangjiang Yang and Ronan Le Bras and Oyvind Tafjord and Chris Wilhelm and Luca Soldaini and Noah A. Smith and Yizhong Wang and Pradeep Dasigi and Hannaneh Hajishirzi},
      year={2025},
      eprint={2411.15124},
      archivePrefix={arXiv},
      primaryClass={cs.CL},
      url={https://arxiv.org/abs/2411.15124}, 
}

@misc{shao2024deepseekmath,
      title={DeepSeekMath: Pushing the Limits of Mathematical Reasoning in Open Language Models}, 
      author={Zhihong Shao and Peiyi Wang and Qihao Zhu and Runxin Xu and Junxiao Song and Xiao Bi and Haowei Zhang and Mingchuan Zhang and Y. K. Li and Y. Wu and Daya Guo},
      year={2024},
      eprint={2402.03300},
      archivePrefix={arXiv},
      primaryClass={cs.CL},
      url={https://arxiv.org/abs/2402.03300}, 
}

@misc{deepseekai2025r1,
      title={DeepSeek-R1: Incentivizing Reasoning Capability in LLMs via Reinforcement Learning}, 
      author={DeepSeek-AI},
      year={2025},
      eprint={2501.12948},
      archivePrefix={arXiv},
      primaryClass={cs.CL},
      url={https://arxiv.org/abs/2501.12948}, 
}

@inproceedings{
yu2025dapo,
title={{DAPO}: An Open-Source {LLM} Reinforcement Learning System at Scale},
author={Qiying Yu and Zheng Zhang and Ruofei Zhu and Yufeng Yuan and Xiaochen Zuo and YuYue and Weinan Dai and Tiantian Fan and Gaohong Liu and Juncai Liu and LingJun Liu and Xin Liu and Haibin Lin and Zhiqi Lin and Bole Ma and Guangming Sheng and Yuxuan Tong and Chi Zhang and Mofan Zhang and Ru Zhang and Wang Zhang and Hang Zhu and Jinhua Zhu and Jiaze Chen and Jiangjie Chen and Chengyi Wang and Hongli Yu and Yuxuan Song and Xiangpeng Wei and Hao Zhou and Jingjing Liu and Wei-Ying Ma and Ya-Qin Zhang and Lin Yan and Yonghui Wu and Mingxuan Wang},
booktitle={The Thirty-ninth Annual Conference on Neural Information Processing Systems},
year={2025},
url={https://openreview.net/forum?id=2a36EMSSTp}
}

@misc{li2025limr,
  author = {Li, Xuefeng and Zou, Haoyang and Liu, Pengfei},
  title = {LIMR: Less is More for RL Scaling},
  year = {2025},
  publisher = {GitHub},
  journal = {GitHub repository},
  howpublished = {\url{https://github.com/GAIR-NLP/LIMR}},
}

@article{wang2025oneshot,
  title={Reinforcement Learning for Reasoning in Large Language Models with One Training Example},
  author={Wang, Yiping and Yang, Qing and Zeng, Zhiyuan and Ren, Liliang and Liu, Lucas and Peng, Baolin and Cheng, Hao and He, Xuehai and Wang, Kuan and Gao, Jianfeng and Chen, Weizhu and Wang, Shuohang and Du, Simon Shaolei and Shen, Yelong},
  journal={arXiv preprint arXiv:2504.20571},
  year={2025}
}

@inproceedings{xia2024less,
   title={{LESS}: Selecting Influential Data for Targeted Instruction Tuning},
   author={Xia, Mengzhou and Malladi, Sadhika and Gururangan, Suchin and Arora, Sanjeev and Chen, Danqi},
   booktitle={International Conference on Machine Learning (ICML)},
   year={2024}
}

@inproceedings{learnalign,
    title = "{L}earn{A}lign: Data Selection for {LLM} Reinforcement Learning with Improved Gradient Alignment",
    author = "Li, Shipeng  and
      Yang, Zhiqin  and
      Li, Shikun  and
      Xia, Xiaobo  and
      Liu, Hengyu  and
      Zhang, Xinghua  and
      Chen, Gaode  and
      Fang, Dong  and
      Tai, Ying  and
      Peng, Zhe",
    editor = "Liakata, Maria  and
      Moreira, Viviane P.  and
      Zhang, Jiajun  and
      Jurgens, David",
    booktitle = "Findings of the {A}ssociation for {C}omputational {L}inguistics: {ACL} 2026",
    month = jul,
    year = "2026",
    address = "San Diego, California, United States",
    publisher = "Association for Computational Linguistics",
    url = "https://aclanthology.org/2026.findings-acl.2009/",
    doi = "10.18653/v1/2026.findings-acl.2009",
    pages = "40417--40434",
    ISBN = "979-8-89176-395-1",

}

@inproceedings{
yang2026gradalign,
title={GradAlign: Gradient-Aligned Data Selection for {LLM} Reinforcement Learning},
author={Ningyuan Yang and Weihua Du and Weiwei Sun and Sean Welleck and Yiming Yang},
booktitle={Third Conference on Language Modeling},
year={2026},
url={https://openreview.net/forum?id=bGUY6CJSbu}
}

@inproceedings{zhouLessismore2023,
 author = {Zhou, Chunting and Liu, Pengfei and Xu, Puxin and Iyer, Srinivasan and Sun, Jiao and Mao, Yuning and Ma, Xuezhe and Efrat, Avia and Yu, Ping and YU, LILI and Zhang, Susan and Ghosh, Gargi and Lewis, Mike and Zettlemoyer, Luke and Levy, Omer},
 booktitle = {Advances in Neural Information Processing Systems},
 doi = {10.52202/075280-2400},
 editor = {A. Oh and T. Naumann and A. Globerson and K. Saenko and M. Hardt and S. Levine},
 pages = {55006--55021},
 publisher = {Curran Associates, Inc.},
 title = {LIMA: Less Is More for Alignment},
 url = {https://proceedings.neurips.cc/paper_files/paper/2023/file/ac662d74829e4407ce1d126477f4a03a-Paper-Conference.pdf},
 volume = {36},
 year = {2023}
}

@inproceedings{
ye2025limo,
title={{LIMO}: Less is More for Reasoning},
author={Yixin Ye and Zhen Huang and Yang Xiao and Ethan Chern and Shijie Xia and Pengfei Liu},
booktitle={Second Conference on Language Modeling},
year={2025},
url={https://openreview.net/forum?id=T2TZ0RY4Zk}
}

@misc{muennighoff2025s1,
      title={s1: Simple test-time scaling}, 
      author={Niklas Muennighoff and Zitong Yang and Weijia Shi and Xiang Lisa Li and Li Fei-Fei and Hannaneh Hajishirzi and Luke Zettlemoyer and Percy Liang and Emmanuel Candès and Tatsunori Hashimoto},
      year={2025},
      eprint={2501.19393},
      archivePrefix={arXiv},
      primaryClass={cs.CL},
      url={https://arxiv.org/abs/2501.19393}, 
}

@inproceedings{sener2018active,
    title={Active Learning for Convolutional Neural Networks: A Core-Set Approach},
    author={Ozan Sener and Silvio Savarese},
    booktitle={International Conference on Learning Representations},
    year={2018},
    url={https://openreview.net/forum?id=H1aIuk-RW},
}

@misc{marion2023when,
      title={When Less is More: Investigating Data Pruning for Pretraining LLMs at Scale}, 
      author={Max Marion and Ahmet Üstün and Luiza Pozzobon and Alex Wang and Marzieh Fadaee and Sara Hooker},
      year={2023},
      eprint={2309.04564},
      archivePrefix={arXiv},
      primaryClass={cs.CL},
      url={https://arxiv.org/abs/2309.04564}, 
}

@inproceedings{li2026answerdivergence,
    title = "Instruction Data Selection via Answer Divergence",
    author = "Li, Bo  and
      Wang, Mingda  and
      Zhang, Shikun  and
      Ye, Wei",
    editor = "Liakata, Maria  and
      Moreira, Viviane P.  and
      Zhang, Jiajun  and
      Jurgens, David",
    booktitle = "Proceedings of the 64th Annual Meeting of the {A}ssociation for {C}omputational {L}inguistics (Volume 1: Long Papers)",
    month = jul,
    year = "2026",
    address = "San Diego, California, United States",
    publisher = "Association for Computational Linguistics",
    url = "https://aclanthology.org/2026.acl-long.214/",
    doi = "10.18653/v1/2026.acl-long.214",
    pages = "4684--4702",
    ISBN = "979-8-89176-390-6",
}

@misc{zheng2025greso,
      title={Act Only When It Pays: Efficient Reinforcement Learning for LLM Reasoning via Selective Rollouts}, 
      author={Haizhong Zheng and Yang Zhou and Brian R. Bartoldson and Bhavya Kailkhura and Fan Lai and Jiawei Zhao and Beidi Chen},
      year={2025},
      eprint={2506.02177},
      archivePrefix={arXiv},
      primaryClass={cs.AI},
      url={https://arxiv.org/abs/2506.02177}, 
}

@inproceedings{bae2026online,
    title = "Online Difficulty Filtering for Reasoning Oriented Reinforcement Learning",
    author = "Bae, Sanghwan  and
      Hong, Jiwoo  and
      Lee, Min Young  and
      Kim, Hanbyul  and
      Nam, Jeongyeon  and
      Kwak, Donghyun",
    editor = "Demberg, Vera  and
      Inui, Kentaro  and
      Marquez, Llu{\'i}s",
    booktitle = "Proceedings of the 19th Conference of the {E}uropean Chapter of the {A}ssociation for {C}omputational {L}inguistics (Volume 1: Long Papers)",
    month = mar,
    year = "2026",
    address = "Rabat, Morocco",
    publisher = "Association for Computational Linguistics",
    url = "https://aclanthology.org/2026.eacl-long.30/",
    doi = "10.18653/v1/2026.eacl-long.30",
    pages = "700--719",
    ISBN = "979-8-89176-380-7",
}

@inproceedings{
    sun2025difficultytargeted,
    title={Improving Data Efficiency for {LLM} Reinforcement Fine-tuning Through Difficulty-targeted Online Data Selection and Rollout Replay},
    author={Yifan Sun and Jingyan Shen and Yibin Wang and Tianyu Chen and Zhendong Wang and Mingyuan Zhou and Huan Zhang},
    booktitle={The Thirty-ninth Annual Conference on Neural Information Processing Systems},
    year={2025},
    url={https://openreview.net/forum?id=uwUkETPIJN}
}

@misc{kim2026spendrollouts,
      title={Spend Your Rollouts Where It Counts: Rollout Allocation for Group-Based RL Post-Training}, 
      author={Woojeong Kim and Ziyi Yang and Jing Nathan Yan and Jialu Liu},
      year={2026},
      eprint={2605.26606},
      archivePrefix={arXiv},
      primaryClass={cs.LG},
      url={https://arxiv.org/abs/2605.26606}, 
}

@inproceedings{
    wu2026singlerollout,
    title={Single-Rollout Hidden-State Dynamics for Training-Free {RLVR} Data Selection},
    author={Jianghao Wu and Jianfei Cai and Weiqiang Wang and Jin Ye and Daniel F. Schmidt and Yasmeen George},
    booktitle={Forty-third International Conference on Machine Learning},
    year={2026},
    url={https://openreview.net/forum?id=pvq2AhjOy1}
}

@article{math500,
      title={Let's Verify Step by Step}, 
      author={Lightman, Hunter and Kosaraju, Vineet and Burda, Yura and Edwards, Harri and Baker, Bowen and Lee, Teddy and Leike, Jan and Schulman, John and Sutskever, Ilya and Cobbe, Karl},
      journal={arXiv preprint arXiv:2305.20050},
      year={2023}
}

@inproceedings{10.1145/3689031.3696075,
author = {Sheng, Guangming and Zhang, Chi and Ye, Zilingfeng and Wu, Xibin and Zhang, Wang and Zhang, Ru and Peng, Yanghua and Lin, Haibin and Wu, Chuan},
title = {HybridFlow: A Flexible and Efficient RLHF Framework},
year = {2025},
isbn = {9798400711961},
publisher = {Association for Computing Machinery},
address = {New York, NY, USA},
url = {https://doi.org/10.1145/3689031.3696075},
doi = {10.1145/3689031.3696075},
booktitle = {Proceedings of the Twentieth European Conference on Computer Systems},
pages = {1279–1297},
numpages = {19},
location = {Rotterdam, Netherlands},
series = {EuroSys '25}
}

@article{yang2024qwen2,
  title={Qwen2 technical report},
  author={Yang, An and Yang, Baosong and Hui, Binyuan and Zheng, Bo and Yu, Bowen and Zhou, Chang and Li, Chengpeng and Li, Chengyuan and Liu, Dayiheng and Huang, Fei and others},
  journal={arXiv preprint arXiv:2407.10671},
  year={2024}
}

@InProceedings{Etha2022Pvi,
  title = 	 {Understanding Dataset Difficulty with $\mathcal{V}$-Usable Information},
  author =       {Ethayarajh, Kawin and Choi, Yejin and Swayamdipta, Swabha},
  booktitle = 	 {Proceedings of the 39th International Conference on Machine Learning},
  pages = 	 {5988--6008},
  year = 	 {2022},
  editor = 	 {Chaudhuri, Kamalika and Jegelka, Stefanie and Song, Le and Szepesvari, Csaba and Niu, Gang and Sabato, Sivan},
  volume = 	 {162},
  series = 	 {Proceedings of Machine Learning Research},
  month = 	 {17--23 Jul},
  publisher =    {PMLR},
  url = 	 {https://proceedings.mlr.press/v162/ethayarajh22a.html},
}

@Misc{mathruler,
  title = {MathRuler},
  author = {hiyouga},
  howpublished = {\url{https://github.com/hiyouga/MathRuler}},
  year = {2025}
}

@misc{aime24,
      title={American Invitational Mathematics Examination (AIME) 2024}, 
      author={Zhang, Yifan and Math-AI, Team},
      year={2024},
}

@inproceedings{Zhang2023gaokao,
  title={Evaluating the Performance of Large Language Models on GAOKAO Benchmark},
  author={Xiaotian Zhang and Chunyang Li and Yi Zong and Zhengyu Ying and Liang He and Xipeng Qiu},
  year={2023}
}

@inproceedings{zhou-etal-2024-paraphrase,
    title = "Paraphrase and Solve: Exploring and Exploiting the Impact of Surface Form on Mathematical Reasoning in Large Language Models",
    author = "Zhou, Yue  and
      Zhu, Yada  and
      Antognini, Diego  and
      Kim, Yoon  and
      Zhang, Yang",
    editor = "Duh, Kevin  and
      Gomez, Helena  and
      Bethard, Steven",
    booktitle = "Proceedings of the 2024 Conference of the North American Chapter of the Association for Computational Linguistics: Human Language Technologies (Volume 1: Long Papers)",
    month = jun,
    year = "2024",
    address = "Mexico City, Mexico",
    publisher = "Association for Computational Linguistics",
    url = "https://aclanthology.org/2024.naacl-long.153/",
    doi = "10.18653/v1/2024.naacl-long.153",
    pages = "2793--2804",
}

@inproceedings{bae-etal-2026-online,
    title = "Online Difficulty Filtering for Reasoning Oriented Reinforcement Learning",
    author = "Bae, Sanghwan  and
      Hong, Jiwoo  and
      Lee, Min Young  and
      Kim, Hanbyul  and
      Nam, Jeongyeon  and
      Kwak, Donghyun",
    editor = "Demberg, Vera  and
      Inui, Kentaro  and
      Marquez, Llu{\'i}s",
    booktitle = "Proceedings of the 19th Conference of the {E}uropean Chapter of the {A}ssociation for {C}omputational {L}inguistics (Volume 1: Long Papers)",
    month = mar,
    year = "2026",
    address = "Rabat, Morocco",
    publisher = "Association for Computational Linguistics",
    url = "https://aclanthology.org/2026.eacl-long.30/",
    doi = "10.18653/v1/2026.eacl-long.30",
    pages = "700--719",
    ISBN = "979-8-89176-380-7",

}

@misc{AMC,
  author       = {{Art of Problem Solving}},
  title        = {AMC Problems and Solutions},
  howpublished = {\url{https://artofproblemsolving.com/wiki/index.php?title=AMC_Problems_and_Solutions}},
  note         = {Accessed: 2025-04-20},
  year={2024}
}

@inproceedings{Minerva,
 author = {Lewkowycz, Aitor and Andreassen, Anders and Dohan, David and Dyer, Ethan and Michalewski, Henryk and Ramasesh, Vinay and Slone, Ambrose and Anil, Cem and Schlag, Imanol and Gutman-Solo, Theo and Wu, Yuhuai and Neyshabur, Behnam and Gur-Ari, Guy and Misra, Vedant},
 booktitle = {Advances in Neural Information Processing Systems},
 doi = {10.52202/068431-0278},
 editor = {S. Koyejo and S. Mohamed and A. Agarwal and D. Belgrave and K. Cho and A. Oh},
 pages = {3843--3857},
 publisher = {Curran Associates, Inc.},
 title = {Solving Quantitative Reasoning Problems with Language Models},
 url = {https://proceedings.neurips.cc/paper_files/paper/2022/file/18abbeef8cfe9203fdf9053c9c4fe191-Paper-Conference.pdf},
 volume = {35},
 year = {2022}
}

@article{he2024olympiadbench,
  title={Olympiadbench: A challenging benchmark for promoting agi with olympiad-level bilingual multimodal scientific problems},
  author={He, Chaoqun and Luo, Renjie and Bai, Yuzhuo and Hu, Shengding and Thai, Zhen Leng and Shen, Junhao and Hu, Jinyi and Han, Xu and Huang, Yujie and Zhang, Yuxiang and others},
  journal={arXiv preprint arXiv:2402.14008},
  year={2024}
}

\appendix

\section{Per-Seed Results Under Greedy Decoding}
\label{app:greedy}

\begin{table*}[!t]
\centering
\footnotesize
\setlength{\tabcolsep}{5pt}
\begin{tabular}{cclccccccc}
\toprule
\multicolumn{3}{c}{\textbf{Training dataset}} & \multicolumn{7}{c}{\textbf{Evaluation benchmarks}} \\
\cmidrule(lr){1-3} \cmidrule(lr){4-10}
Dataset & Budget & Method & Math500 & AIME24 & AMC & Gaokao & Miner. & Olymp. & Avg. \\
\midrule
\multicolumn{10}{c}{\textit{Qwen2.5-Math-1.5B}} \\
\midrule
\multicolumn{3}{c}{Base model} & 40.6 & 10.0 & 37.5 & 17.5 & 7.4 & 23.9 & 22.8 \\
\midrule
  \multirow{10}{*}{\shortstack{MATH\\\scriptsize 350 train}} & \multirow{4}{*}{2\% (7)} & Random & \ms{65.2}{2.7} & \ms{\textbf{10.4}}{5.5} & \ms{\textbf{49.1}}{5.8} & \ms{53.1}{1.8} & \ms{16.5}{1.2} & \ms{\textbf{29.8}}{1.3} & \ms{\textbf{37.4}}{1.8} \\
   &  & SHIFT & \ms{64.7}{2.1} & \ms{6.7}{7.2} & \ms{46.9}{6.9} & \ms{55.2}{2.6} & \ms{17.9}{1.3} & \ms{28.7}{1.5} & \ms{36.7}{1.0} \\
   &  & SHIFT-ES & \ms{\textbf{67.7}}{1.4} & \ms{5.8}{1.7} & \ms{45.6}{5.5} & \ms{\textbf{57.2}}{0.9} & \ms{\textbf{19.5}}{1.2} & \ms{27.9}{0.9} & \ms{37.3}{0.7} \\
   &  & LIFT (ours) & \ms{64.0}{3.7} & \ms{8.3}{1.9} & \ms{41.9}{4.7} & \ms{55.0}{1.6} & \ms{17.4}{1.5} & \ms{29.7}{1.4} & \ms{36.1}{0.6} \\
\cmidrule(lr){2-10}
   & \multirow{4}{*}{4\% (14)} & Random & \ms{65.8}{2.9} & \ms{11.7}{3.6} & \ms{\textbf{51.2}}{3.0} & \ms{55.4}{1.3} & \ms{18.2}{1.2} & \ms{30.2}{1.8} & \ms{38.7}{1.0} \\
   &  & SHIFT & \ms{68.7}{3.7} & \ms{10.8}{4.2} & \ms{47.5}{7.4} & \ms{57.2}{2.7} & \ms{\textbf{19.2}}{3.0} & \ms{30.6}{1.0} & \ms{39.0}{0.7} \\
   &  & SHIFT-ES & \ms{70.5}{3.4} & \ms{5.8}{5.7} & \ms{48.1}{5.2} & \ms{\textbf{58.9}}{2.1} & \ms{\textbf{19.2}}{2.0} & \ms{30.2}{1.9} & \ms{38.8}{0.6} \\
   &  & LIFT (ours) & \ms{\textbf{71.3}}{4.5} & \ms{\textbf{12.5}}{3.2} & \ms{47.5}{4.6} & \ms{58.6}{3.1} & \ms{19.1}{1.9} & \ms{\textbf{31.3}}{1.7} & \ms{\textbf{40.1}}{2.0} \\
\cmidrule(lr){2-10}
   & \multirow{2}{*}{100\%} & Full data & \ms{71.0}{2.6} & \ms{11.7}{3.3} & \ms{51.9}{5.5} & \ms{57.4}{1.4} & \ms{17.6}{2.6} & \ms{31.9}{1.1} & \ms{40.3}{0.5} \\
   &  & GRESO & \ms{71.0}{3.0} & \ms{9.2}{1.7} & \ms{50.0}{5.4} & \ms{58.7}{2.3} & \ms{19.7}{1.8} & \ms{31.8}{1.1} & \ms{40.1}{1.9} \\
\midrule
  \multirow{10}{*}{\shortstack{LIMR\\\scriptsize 1{,}389}} & \multirow{4}{*}{2\% (28)} & Random & \ms{68.7}{1.2} & \ms{12.1}{4.0} & \ms{50.3}{2.5} & \ms{\textbf{57.0}}{1.3} & \ms{20.2}{1.8} & \ms{31.5}{0.8} & \ms{40.0}{0.7} \\
   &  & SHIFT & \ms{\textbf{70.0}}{0.7} & \ms{10.8}{3.2} & \ms{\textbf{55.0}}{6.1} & \ms{56.1}{0.9} & \ms{18.4}{1.6} & \ms{\textbf{32.5}}{1.4} & \ms{40.5}{1.9} \\
   &  & SHIFT-ES & \ms{69.5}{0.4} & \ms{10.0}{5.8} & \ms{54.2}{3.8} & \ms{55.7}{0.8} & \ms{19.9}{0.4} & \ms{30.5}{0.3} & \ms{40.0}{0.5} \\
   &  & LIFT (ours) & \ms{69.9}{1.0} & \ms{\textbf{12.5}}{5.7} & \ms{54.4}{4.3} & \ms{56.8}{1.0} & \ms{\textbf{20.5}}{2.6} & \ms{31.8}{0.1} & \ms{\textbf{41.0}}{1.5} \\
\cmidrule(lr){2-10}
   & \multirow{4}{*}{4\% (56)} & Random & \ms{68.2}{1.0} & \ms{12.9}{2.8} & \ms{52.2}{8.0} & \ms{56.3}{1.1} & \ms{18.7}{0.8} & \ms{31.2}{1.3} & \ms{39.9}{1.6} \\
   &  & SHIFT & \ms{\textbf{71.0}}{0.7} & \ms{10.8}{6.3} & \ms{54.4}{6.9} & \ms{56.5}{0.3} & \ms{20.1}{1.2} & \ms{31.5}{0.4} & \ms{40.7}{1.6} \\
   &  & SHIFT-ES & \ms{70.9}{0.8} & \ms{13.3}{2.7} & \ms{51.2}{4.3} & \ms{56.7}{0.6} & \ms{18.9}{1.5} & \ms{31.8}{0.5} & \ms{40.5}{1.0} \\
   &  & LIFT (ours) & \ms{70.0}{1.6} & \ms{\textbf{16.7}}{2.7} & \ms{\textbf{56.2}}{5.2} & \ms{\textbf{57.3}}{1.2} & \ms{\textbf{20.4}}{0.8} & \ms{\textbf{32.0}}{0.3} & \ms{\textbf{42.1}}{0.9} \\
\cmidrule(lr){2-10}
   & \multirow{2}{*}{100\%} & Full data & \ms{70.1}{1.1} & \ms{14.2}{1.7} & \ms{58.8}{3.2} & \ms{55.5}{1.2} & \ms{19.9}{2.1} & \ms{32.3}{0.7} & \ms{41.8}{0.2} \\
   &  & GRESO & \ms{70.8}{0.6} & \ms{12.5}{5.0} & \ms{54.4}{2.4} & \ms{57.0}{0.9} & \ms{20.7}{1.5} & \ms{32.2}{1.1} & \ms{41.3}{1.1} \\
\midrule
\multicolumn{10}{c}{\textit{DeepSeek-R1-Distill-Qwen-1.5B}} \\
\midrule
\multicolumn{3}{c}{Base model} & 67.2 & 13.3 & 47.5 & 58.0 & 17.6 & 28.5 & 38.7 \\
\midrule
  \multirow{10}{*}{\shortstack{LIMR\\\scriptsize 1{,}389}} & \multirow{4}{*}{2\% (28)} & Random & \ms{69.2}{2.8} & \ms{15.8}{6.3} & \ms{51.2}{6.0} & \ms{60.6}{3.5} & \ms{20.1}{1.3} & \ms{31.2}{1.9} & \ms{41.4}{2.0} \\
   &  & SHIFT & \ms{71.5}{3.5} & \ms{18.3}{7.9} & \ms{\textbf{56.9}}{5.5} & \ms{61.8}{5.0} & \ms{21.5}{3.1} & \ms{32.6}{4.7} & \ms{43.8}{3.7} \\
   &  & SHIFT-ES & \ms{\textbf{73.5}}{3.6} & \ms{20.0}{8.8} & \ms{55.8}{11.8} & \ms{62.8}{7.6} & \ms{\textbf{22.5}}{4.9} & \ms{\textbf{33.9}}{4.4} & \ms{\textbf{44.7}}{6.7} \\
   &  & LIFT (ours) & \ms{72.2}{3.5} & \ms{\textbf{20.8}}{5.0} & \ms{\textbf{56.9}}{5.5} & \ms{\textbf{62.8}}{3.7} & \ms{21.2}{3.2} & \ms{32.5}{3.4} & \ms{44.4}{2.9} \\
\cmidrule(lr){2-10}
   & \multirow{4}{*}{4\% (56)} & Random & \ms{73.3}{4.7} & \ms{\textbf{21.0}}{1.6} & \ms{\textbf{56.4}}{5.9} & \ms{62.3}{4.9} & \ms{21.0}{3.6} & \ms{33.8}{3.7} & \ms{44.6}{3.6} \\
   &  & SHIFT & \ms{\textbf{73.6}}{4.5} & \ms{20.8}{1.7} & \ms{55.0}{8.9} & \ms{\textbf{63.9}}{4.0} & \ms{\textbf{23.3}}{2.6} & \ms{\textbf{34.2}}{3.9} & \ms{\textbf{45.1}}{3.7} \\
   &  & SHIFT-ES & \ms{63.0}{11.4} & \ms{17.8}{6.9} & \ms{55.8}{18.9} & \ms{54.0}{11.0} & \ms{16.7}{4.8} & \ms{28.6}{8.0} & \ms{39.3}{9.6} \\
   &  & LIFT (ours) & \ms{69.9}{1.7} & \ms{20.0}{4.7} & \ms{45.6}{9.4} & \ms{59.9}{3.2} & \ms{18.9}{1.5} & \ms{31.7}{2.1} & \ms{41.0}{2.7} \\
\cmidrule(lr){2-10}
   & \multirow{2}{*}{100\%} & Full data & \ms{73.2}{2.1} & \ms{23.3}{9.0} & \ms{60.6}{8.0} & \ms{63.3}{1.6} & \ms{21.4}{1.5} & \ms{34.6}{2.2} & \ms{46.1}{2.7} \\
   &  & GRESO & \ms{68.4}{9.6} & \ms{15.0}{7.1} & \ms{53.8}{12.4} & \ms{59.0}{7.4} & \ms{21.3}{1.0} & \ms{31.8}{6.0} & \ms{41.6}{7.2} \\
\bottomrule
\end{tabular}
\caption{Greedy pass@1 (\%) for the checkpoints of Table~\ref{tab:main}: one sample per problem at temperature $0$ and top-$p$ $1$, with a $2{,}688$ token cap for Qwen2.5-Math and $12{,}288$ for DeepSeek-R1-Distill. Rows, methods and training seeds match Table~\ref{tab:main} exactly; the selection cost column is omitted because it does not depend on the decoding rule used at evaluation. Bold marks the best method within each budget block.}
\label{tab:greedy}
\end{table*}

Table~\ref{tab:main} reports accuracy under sampled decoding, four samples per problem at temperature $0.6$ and top-$p$ $0.95$.
Table~\ref{tab:greedy} repeats it under greedy decoding, one sample per problem at temperature $0$ and top-$p$ $1$, for the same checkpoints and the same training seeds, with the same layout.

For Qwen2.5-Math the two decoding rules agree closely: every greedy average lies within $1.7$ points of its sampled counterpart, and LIFT has the highest greedy average in three of the four Qwen blocks.
For DeepSeek-R1-Distill they diverge sharply, by $12$ to $18$ points, and the seed spread grows from $0.4$--$1.0$ under sampling to $2.7$--$9.6$ under greedy decoding.
Greedy decoding drives this model into the non-terminating repetitions analysed in Section~\ref{sec:token-prcl}, so a single greedy sample is a poor measure of its quality, which is why Table~\ref{tab:main} reports sampled accuracy for both models.

\section{Prompt Templates}
\label{app:prompts}

Every prompt is built through the tokenizer's chat template, so the conditional context $\mathcal{T}(x)$ and the null context $\mathcal{T}(\varnothing)$ of Eq.~\eqref{eq:st} differ only in the problem text.
The same template is used for selection, for RL training and for evaluation of a given model, so the scored policy and the trained policy see identical inputs.
Table~\ref{tab:prompts} gives both templates and both null contexts.

\begin{table*}[h]
\centering
\begin{tcolorbox}[
  enhanced,
  sharp corners,
  colback=gray!8,
  colframe=black!20,
  boxrule=0.5pt,
  width=1\textwidth,
  title={\textbf{Prompt Templates}}
]
\footnotesize
\textbf{Qwen2.5-Math-1.5B, conditional context $\mathcal{T}(x)$.} \\
\textit{system:} ``You FIRST think about the reasoning process as an internal monologue and then provide the final answer. \\
The reasoning process MUST BE enclosed within \texttt{<think>} \texttt{</think>} tags. \\
The final answer MUST BE put in \textbackslash boxed\{\}.'' \\
\textit{user:} ``\{problem\}'' \\[4pt]

\textbf{Qwen2.5-Math-1.5B, null context $\mathcal{T}(\varnothing)$.} \\
\textit{system:} the same message as above. \\
\textit{user:} the turn is removed entirely, so the problem is the only difference from $\mathcal{T}(x)$. \\[4pt]

\textbf{DeepSeek-R1-Distill-Qwen-1.5B, conditional context $\mathcal{T}(x)$.} \\
\textit{system:} none, as the model's release recommends. \\
\textit{user:} ``\{problem\} \\
Please reason step by step, and put your final answer within \textbackslash boxed\{\}.'' \\[4pt]

\textbf{DeepSeek-R1-Distill-Qwen-1.5B, null context $\mathcal{T}(\varnothing)$.} \\
\textit{system:} none. \\
\textit{user:} ``Please reason step by step, and put your final answer within \textbackslash boxed\{\}.'' \\
The instruction is kept and the problem removed, the analogue of dropping the user turn under the Qwen2.5-Math template.
\end{tcolorbox}
\caption{Prompt templates and null contexts for the two models. Benchmark problems are substituted into the same template as the model's own training prompt.}
\label{tab:prompts}
\end{table*}

\paragraph{Evaluation.}
Benchmark problems are substituted into the same template as the model's own training prompt.
Table~\ref{tab:main} uses four samples per problem at temperature $0.6$ and top-$p$ $0.95$, with a generation cap of $2{,}688$ tokens for Qwen2.5-Math and $32{,}768$ for DeepSeek-R1-Distill.
Answers are extracted from the final \texttt{\textbackslash boxed\{\}} expression and checked for mathematical equivalence against the reference.

\section{What LIFT Responds To}
\label{app:examples}

\paragraph{Two selected prompts.}
Table~\ref{tab:lift-prompts} shows two prompts from LIFT's $2\%$ LIMR selection, one per model, both near the top of their selection by $\mathrm{PRCL}_{\hat{K}}$ and of almost the same length.
They differ in difficulty: the Qwen2.5-Math pick is solved by none of 16 sampled rollouts, while the R1-Distill pick is solved in 7 of 16, close to that model's dataset mean.

\begin{table*}[h]
\centering
\begin{tcolorbox}[
  enhanced,
  sharp corners,
  colback=gray!8,
  colframe=black!20,
  boxrule=0.5pt,
  width=1\textwidth,
  title={\textbf{Prompts Selected by LIFT at the $2\%$ LIMR Budget}}
]
\footnotesize
\textbf{Qwen2.5-Math-1.5B, highest-scoring pick.} $\mathrm{PRCL}_{\hat{K}} = 208.8$, highest of the 28 selected; 207 prompt tokens; solve rate $0.00$ against a dataset mean of $0.33$. \\
\textit{``Dr.\ Jones lives in a country with a progressive tax system. That is, he does not pay any taxes on the first \$20{,}000 in income he makes, he pays $5\%$ taxes on the next \$25{,}000, he pays $10\%$ taxes on the next \$35{,}000, he pays $15\%$ on the next \$50{,}000, and he pays $20\%$ on every additional dollar onwards. If Dr.\ Jones pays \$10{,}000 in taxes, then how much income does he make?''} \\[6pt]

\textbf{DeepSeek-R1-Distill-Qwen-1.5B, second-highest pick.} $\mathrm{PRCL}_{\hat{K}} = 228.7$; 211 prompt tokens; solve rate $0.44$ against a dataset mean of $0.47$. \\
\textit{``Tina the tourist goes on a trip. She starts at the origin and drives north for $10$ units. Then she turns east and as she is turning her camera flies out the window and lands exactly at $(0,10)$. She then drives $9$ units east, turns and drives $8$ units north. She continues this pattern of turning and driving one unit less than after the previous turn, until stopping after driving $1$ unit east. She reaches for her camera only to find it missing. She activates the GPS homing device on her camera and drives back to it in a straight line. What is the equation of this line?''}
\end{tcolorbox}
\caption{Two prompts selected by LIFT, one per model, both near the top of their selection by $\mathrm{PRCL}_{\hat{K}}$ and of almost the same length. Only the first is difficult for the model that selected it.}
\label{tab:lift-prompts}
\end{table*}

Both are long, multi-step and narratively framed, and both score near the top of their selection. Only the first is difficult.
Averaged over the whole selection, LIFT picks prompts of $185$ tokens against a dataset mean of $136$ for Qwen2.5-Math, $262$ against $99$ for R1-Distill, and $362$ against $131$ on the MATH training split.

\paragraph{Length or informativeness.}
The two examples raise the question of whether $\mathrm{PRCL}$ measures how informative a prompt is or simply how long it is, since a longer question is a larger context to remove.
The two are separable, and they separate differently for the two models.
For Qwen2.5-Math the rank correlation between $\mathrm{PRCL}_{\hat{K}}$ and prompt length is $0.41$, and the association between the score and difficulty survives when length is regressed out, falling only from $-0.22$ to $-0.18$.
For R1-Distill the correlation with length is $0.77$ and the association with difficulty does not survive, with a residual rank correlation of $+0.01$.
On that model the difficulty of the selected subsets is therefore attributable to prompt length rather than to additional signal in the score, which is consistent with Section~\ref{sec:solve-rate}, where SHIFT selects markedly harder prompts than LIFT for R1-Distill while the ordering reverses for Qwen2.5-Math.

\end{document}